\documentclass[sigconf, nonacm=true]{acmart}

\setcopyright{none}
\renewcommand\footnotetextcopyrightpermission[1]{}
\usepackage{multirow}
\usepackage{hyperref}

\usepackage[utf8]{inputenc}
\usepackage{amsmath}

\AtBeginDocument{%
  \providecommand\BibTeX{{%
    \normalfont B\kern-0.5em{\scshape i\kern-0.25em b}\kern-0.8em\TeX}}}

\setcopyright{acmcopyright}
\copyrightyear{2021}
\acmYear{2021}
\acmDOI{000000000}

\acmConference[]

\begin{document}

\title{PlumeCityNet: Multi-Resolution Air Quality Forecasting}







\author{Thibaut Cassard}
\affiliation{\institution{Plume Labs}}
\email{thibaut.cassard@plumelabs.com}

\author{Grégoire Jauvion}
\affiliation{\institution{Plume Labs}}
\email{gregoire.jauvion@plumelabs.com}

\author{Antoine Alléon}
\affiliation{\institution{Plume Labs}}
\email{antoine.alleon@plumelabs.com}

\author{Boris Quennehen}
\affiliation{\institution{Plume Labs}}
\email{boris.quennehen@plumelabs.com}

\author{David Lissmyr}
\affiliation{\institution{Plume Labs}}
\email{david.lissmyr@plumelabs.com}


\begin{abstract}

This paper presents an engine able to forecast jointly the concentrations of the main pollutants harming people's health: nitrogen dioxide (NO$_{2}$), ozone (O$_{3}$) and particulate matter (PM$_{2.5}$ and PM$_{10}$, which are respectively the particles whose diameters are below $2.5~\mu m$ and $10~\mu m$ respectively).
The engine is fed with air quality monitoring stations' measurements, weather forecasts, physical models' outputs and traffic estimates to produce forecasts up to $24$ hours.
The forecasts are produced with several spatial resolutions, from a few dozens of meters to dozens of kilometers, fitting several use-cases needing air quality data.

We introduce the \textit{Scale-Unit} block, which enables to integrate seamlessly all available inputs at a given resolution to return forecasts at the same resolution.
Then, the engine is based on a U-Net architecture built with several of those blocks, giving it the ability to process inputs and to output predictions at different resolutions.

We have implemented and evaluated the engine on the largest cities in Europe and the United States, and it clearly outperforms other prediction methods.
In particular, the out-of-sample accuracy remains high, meaning that the engine can be used in cities which are not included in the training dataset.
A valuable advantage of the engine is that it does not need much computing power: the forecasts can be built in a few minutes on a standard CPU\@.
Thus, they can be updated very frequently, as soon as new air quality monitoring stations' measurements are available (generally every hour), which is not the case of physical models traditionally used for air quality forecasting.

\end{abstract}




\maketitle

\section{Introduction}
\label{sec:introduction}

Air pollution is one of the major public health concerns.
The World Health Organization (WHO) estimates that more than 80\% of citizens living in urban environments where air quality is monitored are exposed to air quality levels that exceed WHO guideline limits.
It also estimates that $4.2$ million deaths every year are linked to outdoor air pollution exposure~\cite{WHO_report}.
Despite those alarming figures, very few citizens have access to information about the quality of the air they breathe.
More and more public and private initiatives are being developed to close this gap and give to citizens the information they need to protect themselves from air pollution.

A key difficulty to be able to provide relevant air quality information is the lack of data.
Indeed, it is believed that there are about $30$ thousands of air quality monitoring stations worldwide, which is orders of magnitude less than the number of stations needed given the spatial variability of air quality.
For this reason, it is necessary to build air quality models giving hyperlocal air quality forecasts based on the available data.

Air quality modeling is particularly challenging because air quality varies a lot, both in time and in space.
For example, a polluted air can become clean in a few hours after a heavy rain, showing the high temporal variability of air pollution.
Regarding spatial variability, a very important characteristic of air pollution is that its spatial distribution comes from several phenomena happening at different scales~\cite{AQ_review}.
For example, a street crowded with vehicles may be $5$ or $10$ times more polluted than a green park a few hundred meters away.
On a much lower spatial resolution, a significant source of emissions (e.g. a coal power plant) may impact air quality a few thousands of kilometers away on timescales of several days or weeks.
An accurate air quality prediction engine must model those phenomena happening at different scales, and the existing scientific literature falls short on this point.

Also, different applications fed with air quality data may have very different requirements, highlighting the need for air quality data at several spatial and temporal scales.
For example, building a routing engine for pedestrians suggesting itineraries with a low air pollution exposure needs to be fed with data with the highest possible resolution.
However, if one's goal is to rank cities based on their average air quality levels, an air quality prediction with a resolution on the order of a kilometer is much more useful.

This paper presents a forecasting engine able to produce $24$-hour air quality forecasts with different spatial resolutions, from $50$ meters to dozens of kilometers.
The engine covers the main atmospheric pollutants harming people's health and regulated by WHO: nitrogen dioxide (NO$_{2}$), ozone (O$_{3}$) and particulate matter (PM$_{2.5}$ and PM$_{10}$, which are respectively the particles whose diameters are below $2.5~\mu m$ and $10~\mu m$ respectively).
It is fed with the air quality measurements provided by thousands of monitoring stations, weather forecasts, physical models' outputs and traffic estimates.
It is trained and evaluated on the largest cities in Europe and the United States.

The building block of the engine is the \textit{Scale-Unit} block, a powerful architecture based on the encoder-decoder framework and the Convolutional LSTM layer, first introduced in \cite{plumenet}, able to integrate seamlessly the different kinds of inputs fed to the model: historical features (e.g.\ historical traffic estimates), constant features (e.g. a description of the road network) and forecast features (e.g. weather forecasts).

Then, the engine is based on a U-Net architecture formed with stacked \textit{Scale-Unit} blocks, a natural choice in order to build a multi-resolution model.
Basically, every input fed to the engine can be processed with a specific resolution whose choice is driven by how local is its impact on air quality.
Also, the air quality forecasts can be built with several resolutions as well, which is needed to give a comprehensive understanding of air quality over the area.
The engine implemented in this paper inputs features defined at $2$ spatial resolutions, $50$ meters and $20$ kilometers, and the  forecasts are produced with $4$ resolutions, from $50$ meters to $20$ kilometers.

The paper is organized as follows.
We discuss earlier works in Section 2.
Section 3 gives a detailed overview of the data sources used by the engine.
Section 4 presents the architecture of the forecasting engine and details the model estimation process.
Section 5 provides an evaluation of the forecasting engine.

\section{Related work}
\label{sec:related-work}

The problem of air quality prediction is much studied in the literature and is tackled through various angles. \cite{bigdata_summary} and \cite{deep_learning_review} present comprehensive reviews of air quality modeling using machine learning approaches.

\cite{deepplume}, \cite{global_approach} and \cite{r_package_airpred} focus on spatial modeling and do not take into account air pollution temporal variability.
They model the main air pollutants spatial variability at a given time using diverse datasets including monitoring stations measurements, satellite-based measurements, land-use datasets and traffic datasets.
They reach very fine resolutions, from 10 meters in \cite{deepplume} to a few kilometers in \cite{global_approach} and \cite{r_package_airpred}.

Other papers focus on air quality temporal variability and aim at predicting air pollutants future concentrations at a given location.
A commonly used framework consists in building air quality forecasts at a given monitoring station using the station's past measurements as well as weather forecasts at the station's location.
The models are learnt on historical datasets, and most of recent papers using this approach are based on neural networks, and more specifically on LSTM architectures.
\cite{deepair} builds air quality forecasts in Beijing up to 10 hours using an encoder-decoder LSTM architecture.
\cite{deep_air_net} and \cite{india_lstm_rnn} apply slightly different recurrent architectures in a few Indian cities.
\cite{high_air} uses a hierarchical graph neural network-based forecasting method modeling air pollution at different spatial scales simultaneously with a city-level graph and station-level graphs.
The model is evaluated on $10$ cities in China.

More similarly to the approach we use, several papers use spatiotemporal modeling frameworks to take into account both spatial and temporal variability.
\cite{airnet} introduces a dataset with air quality and meteorological data on a $0.25^{\circ}$ regular grid in China over $2$ years.
\cite{uncertainty} uses a Convolutional LSTM network trained on low-cost sensors' measurements at a very fine spatial and temporal scale in a very limited area of less than $1 km^{2}$.
\cite{conv_lstm_2}, \cite{deep_air_learning}, \cite{deep_air_quality}, \cite{fine-grained}, \cite{spatial_temporal}, \cite{fine_grained} and \cite{deep_distributed} build air quality forecasts at China's monitoring stations using the stations' past measurements and weather forecasts.
Spatial correlations between the stations are included in a deep learning architecture, and the temporal variability is modeled with LSTM layers in \cite{conv_lstm_2}, \cite{deep_air_learning} and~\cite{spatial_temporal}. \cite{u-air} uses a similar approach but integrates land-use and traffic features in the forecasts.
In~\cite{conv_lstm}, the authors build air quality forecasts on a regular grid in the Beijing area and in the London area using a convolutional LSTM architecture similar to the one we use in this paper.
In this paper as well, air quality forecasts are based on past measurements and weather forecasts. \cite{att_conv_lstm} uses an architecture with both convolutional LSTM and attention mechanisms.
In order to limit over-fitting due to the low number of monitoring stations available, other papers like \cite{physical_model} and \cite{model_data_driven} introduce physical modeling of the pollutants' dispersion in the spatiotemporal model.

In~\cite{precipitation_nowcasting} and~\cite{metnet}, the authors build precipitation forecasts with a very fine resolution ($1$ kilometer) over the United States using very similar spatiotemporal architectures:~\cite{precipitation_nowcasting} uses a U-net network and~\cite{metnet} uses attention mechanisms.
The convolutional LSTM block is introduced in~\cite{conv_lstm_theory} and applied on precipitation forecasting.

In~\cite{plumenet} an encoder-decoder architecture based on several convolutional LSTM blocks is used to predict air quality on a $0.5^\circ$-resolution grid over Europe and the United States.
This model performs well to predict smooth weighted average of official monitoring stations' measurements up to 4 days.
This type of architecture is used in this paper as an element (named Scale-Unit) of the final model.

The U-Net architecture was introduced in~\cite{u_net} to perform accurate biomedical image segmentation with limited data and in~\cite{mcnn}, the authors proposed a multi-resolution convolutional neural network architecture to better capture high frequency patterns for inverse problems using U-Net architectures.
In this latter architecture the target is estimated at all scales of the U-shaped model.
The prediction engine presented here derives from these two approaches.

\section{Data sources}
\label{sec:data-sources}

This section details the data sources used by the forecasting engine.

\subsection{Air quality monitoring stations measurements}
\label{subsec:air-quality-monitoring-station-measurements}

\begin{figure}[h]
  \centering
  \includegraphics[width=\linewidth]{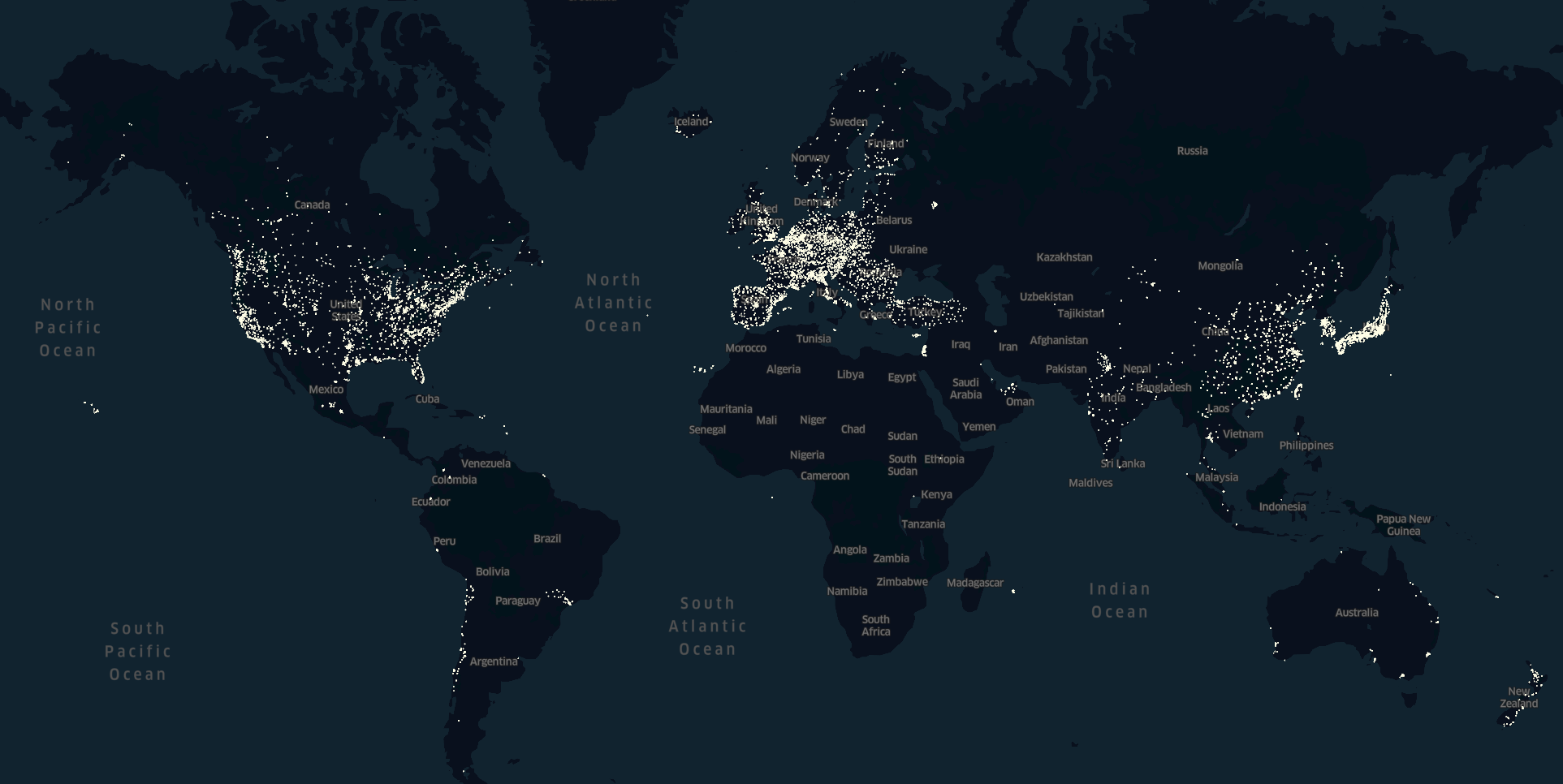}
  \caption{Map of the monitoring stations whose measurements are included in the predictions}
  \label{fig:stations}
\end{figure}

We have built a proprietary architecture based on several dozens of crawlers collecting the air quality measurements provided by about $14000$ monitoring stations across the world.
Figure~\ref{fig:stations} shows a global map of the locations of those monitoring stations.

Almost all monitoring stations measurements are available on an hourly basis.
Each monitoring station does not necessarily measure the four pollutants predicted by the engine.
The experiments presented in this paper have been performed in Europe and the United States, which are the $2$ regions where the number of collected measurements is the highest.
Table~\ref{tab:stations} gives the number of monitoring stations as well as the number of stations measuring each pollutant in those regions.

\begin{table}
  \caption{Number of monitoring stations per region and pollutant}
  \label{tab:stations}
  \begin{tabular}{c||c|cccc}
    \hline
    Region & Global & NO2 & O3 & PM2.5 & PM10 \\
    \hline
    Europe & 2778 & 2252 & 1614 & 790 & 1958 \\
    United States & 1924 & 331 & 1300 & 1041 & 347 \\
    \hline
\end{tabular}
\end{table}

The dataset consists in all measurements from January 1st, 2020 to December 31st, 2020 in a hundred of the largest cities in Europe and the United States.
This represents 2500 monitoring stations with approximately 6.5 million air quality measurements.
We have found that there are missing and erroneous values (generally abnormally high) coming from those monitoring stations.
They can be encountered during station maintenance windows, during station failures or if issues arise during the publishing or collection of said data.
While we are not able to determine the exact cause of such errors, it is important to detect them and define an appropriate treatment: missing values and clearly erroneous values are discarded from the datasets.

\subsection{Traffic Data}
\label{subsec:traffic-data}
Traffic emissions are estimated using two data sources: real-time traffic estimates and a description of the road network.
\subsubsection{Real-time traffic estimates}
\label{sec:traffic}
We collect traffic estimates on a hourly basis across every road segment in Europe and the United States through three features: \textit{JamFactor} (a value between 0 and 10 measuring the road congestion), \textit{Speed} (the estimated speed on the road segment) and \textit{HistoricalSpeed} (a quantile of the speed on the road segment estimated on historical data).
For every hour and cell of a regular grid over the area of interest, we define the feature $\textit{Traffic}$ as a three-dimensional vector whose values are $(\textit{JamFactor}, \textit{Speed}, \textit{HistoricalSpeed})$, being the average value of the features computed on the road segments that intersect the grid cell.
When no road segment intersects the road, $\textit{Traffic} = (0, 0, 0)$.
\subsubsection{Road network}
\label{sec:roads}
Road network details and topology were collected in the regions covered by the prediction engine.
Road network data is collected as a set of road segments, any of them being associated with a set of metadata regrouping a significant amount of information including a classification per usage.
Based on this metadata, road segments are mapped to two aggregate categories named $\textit{Roads}$ and $\textit{MajorRoads}$.
For any cell of a regular grid covering the area of interest, the $\textit{RoadNetwork}$ feature is defined as the two-dimensional vector ($\textit{Roads}$, $\textit{MajorRoads}$) giving the number of road segments mapped to the $\textit{Roads}$ and $\textit{MajorRoads}$ categories that intersect the cell.
\subsection{Weather forecasts}
\label{subsec:weather-forecasts}

Weather simulations are computed using physical and chemical modeling of natural (atmospheric and land-soil) phenomenon.
The weather forecasts provide numerous features like temperature, wind, precipitation, soil moisture or snow depth.
In the experiments presented in this paper, the following features were used: temperature, relative humidity, wind speed and direction (encoded in $u$ and $v$ which are respectively the wind zonal and meridional velocities), planetary boundary layer height and precipitation rate.

The weather forecasts used here are produced daily on a regular grid covering the whole world with a surface resolution of $0.25^{\circ}$ and a hourly time resolution up to the $24$ hours horizon considered here.
We have collected them on a daily basis from January 1st, 2020 to December 31st, 2020.

\subsection{Physical models' outputs}
\label{subsec:aqpcm-outputs}

Physical models rely on physical and chemical modeling of pollutants' emissions, chemical reactions and dispersion, and the simulations are initiated with monitoring stations and satellite-based measurements.
They are characterised by their geographical coverage (regional or global) and their spatial granularity (from a few kilometers to dozens of kilometres).

The forecasts used in this paper have a $0.4^{\circ}$ spatial resolution and a hourly time resolution up to $24$ hours.
They cover the $4$ pollutants forecast by the engine (NO$_{2}$, O$_{3}$, PM$_{2.5}$ and PM$_{10}$) and have been collected on a daily basis from January 1st, 2020 to December 31st, 2020.

\section{Description of the forecasting engine}
\label{sec:description-of-the-forecasting-engine}

The prediction engine is based on a multi-resolution architecture, in the sense that the different input data sources can be processed with different resolutions, and the predictions are produced with several resolutions as well.

In this section, we note $N_{pol}=4$ the number of pollutants forecast by the engine, $N_{in}$ the number of historical time steps and $N_{out}$ the number of forecast time steps.

We first introduce the \textit{Scale-Unit} block, which is the building block of the engine's architecture.
Then, we describe the whole architecture, derived from the well-known U-Net architecture, and we finally describe how it is trained.

\subsection{Introduction of the \textit{Scale-Unit} block}

The \textit{Scale-Unit} block is based on the two following concepts commonly used in spatiotemporal deep learning: the convolutional LSTM layer and the encoder-decoder framework.
It is very similar to the architecture introduced in~\cite{plumenet}.

Convolutional LSTM (ConvLSTM hereafter) is a powerful block to model spatiotemporal data with strong correlations in space.
A ConvLSTM determines the future state of a given grid cell by using the inputs and past states of its local neighbors: this is achieved by using a convolution operator in the state-to-state and input-to-state transitions.
This enables to reduce very significantly the number of parameters of the block compared to a fully-connected LSTM block where every grid cell would be connected to all inputs and past states on the whole grid.
An important feature of the ConvLSTM block is that its number of parameters does not depend on the size of the spatial grid but only on the number of hidden states and on the size of the convolution kernels.
Refer to~\cite{conv_lstm_theory} for a detailed description of the ConvLSTM block.

The encoder-decoder framework has been first introduced for language translation.
In a nutshell, each input feature is encoded onto a grid summarizing the information it contains.
The encoders' outputs are then merged and processed by a decoder for build the predictions, in our case the air quality forecasts at every forecast time step.

\begin{figure*}[h!]
  \centering
  \includegraphics[width=\linewidth]{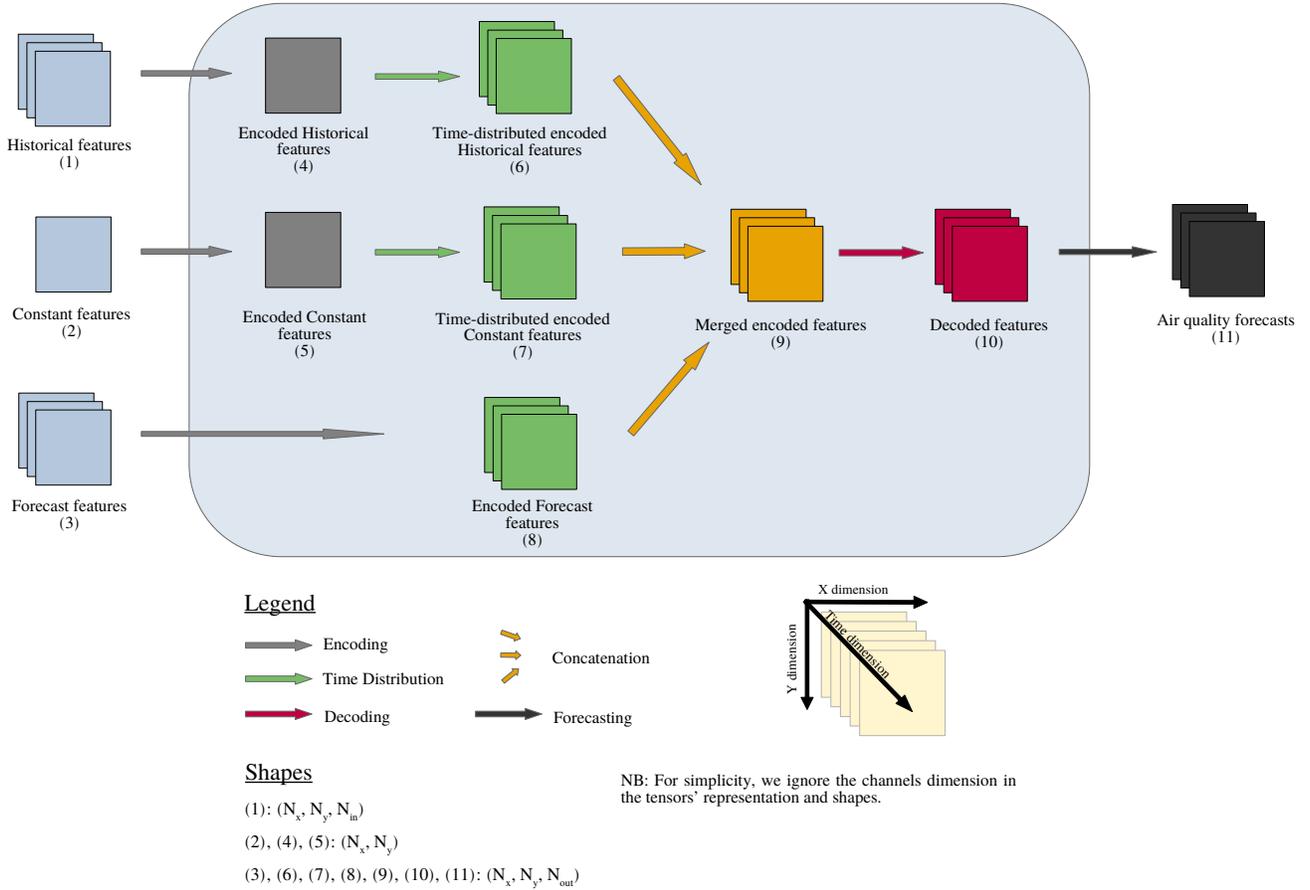}
  \caption{Scale-Unit architecture}
  \label{fig:scale_unit}
\end{figure*}

The \textit{Scale-Unit} block is an implementation of the encoder-decoder framework which takes as inputs historical features (e.g.\ stations' measurements), constant features (e.g.\ road network description) and forecast features (e.g.\ weather forecasts), and which outputs the air quality forecasts on a grid.
The block is schematized on Figure~\ref{fig:scale_unit}.
It is parameterized by the specification of the three encoders (for the historical, constant and forecast features), of the decoder and of the forecasting layer.
We give here their main characteristics and refer to~\cite{plumenet} for more details:
\begin{itemize}
  \item The encoders are formed with stacked ConvLSTM layers or time-distributed 2D convolutional layers.
  The former should be applied to capture spatiotemporal correlations, while the latter is adapted to capture spatial correlations only
  \item The decoder inputs the concatenation of the encoded historical, constant and forecast features, and outputs states which are then fed to the forecasting layer
  \item At every of the $N_{out}$ forecast time steps, the forecasting layer inputs the states produced by the decoder and outputs the air quality forecasts.
  It is formed with a simple 2D convolutional layer with $N_{pol}$ $(1,1)$ kernels and a relu activation function
\end{itemize}

As commonly done to improve convergence when training the model, a batch normalization layer is used after each TimeDistributedConv2D and ConvLSTM block.

\subsection{Architecture of the engine}
\label{subsec:architecture-of-the-engine}

The engine's architecture is inspired by the U-Net architecture built with several stacked \textit{Scale-Unit} blocks.
U-Net is a convolutional neural network architecture introduced in~\cite{u_net} which consists of a contracting path and an expansive path, which gives it a u-shaped architecture.
The contracting path is a typical convolutional network that consists of repeated application of convolutions and pooling operation.
During the contraction, the spatial information is reduced while feature information is increased.
The expansive pathway combines the feature and spatial information through a sequence of up-sampling and concatenations with high-resolution features from the contracting path.

\begin{figure}[h!]
  \centering
  \includegraphics[width=\linewidth]{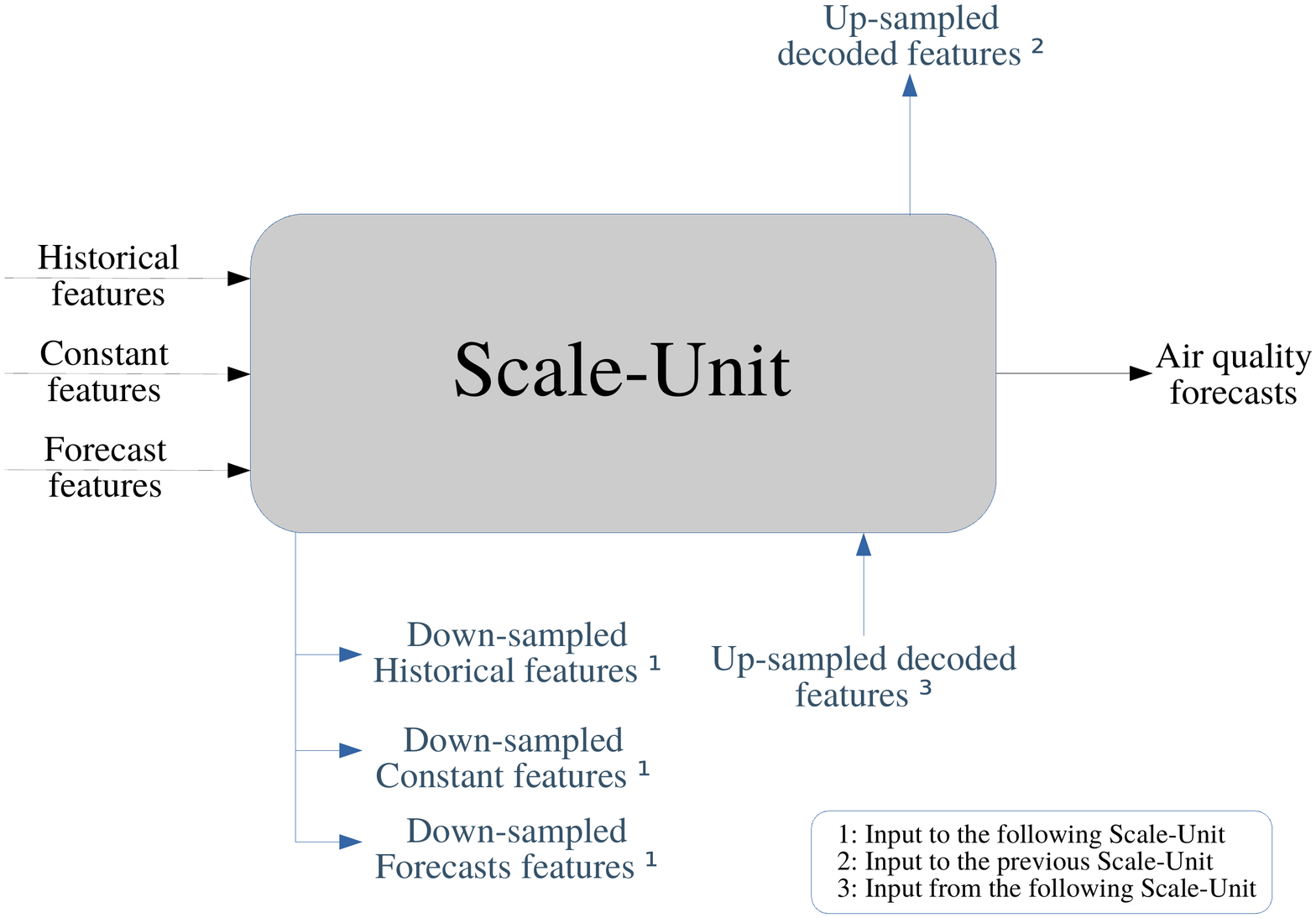}
  \caption{Simplified diagram of the Scale-Unit architecture}
  \label{fig:simplified_scale_unit}
\end{figure}

Figure~\ref{fig:simplified_scale_unit} illustrates how a U-Net architecture can be built by stacking several \textit{Scale-Unit} blocks.
Each one of those blocks is defined by its resolution.
It may also input features decoded by the following \textit{Scale-Unit} block defined by a lower resolution (which are up-sampled using a simple nearest neighbours interpolation), and output its decoded features to the previous \textit{Scale-Unit} block.

\begin{figure}[h!]
  \centering
  \includegraphics[width=\columnwidth]{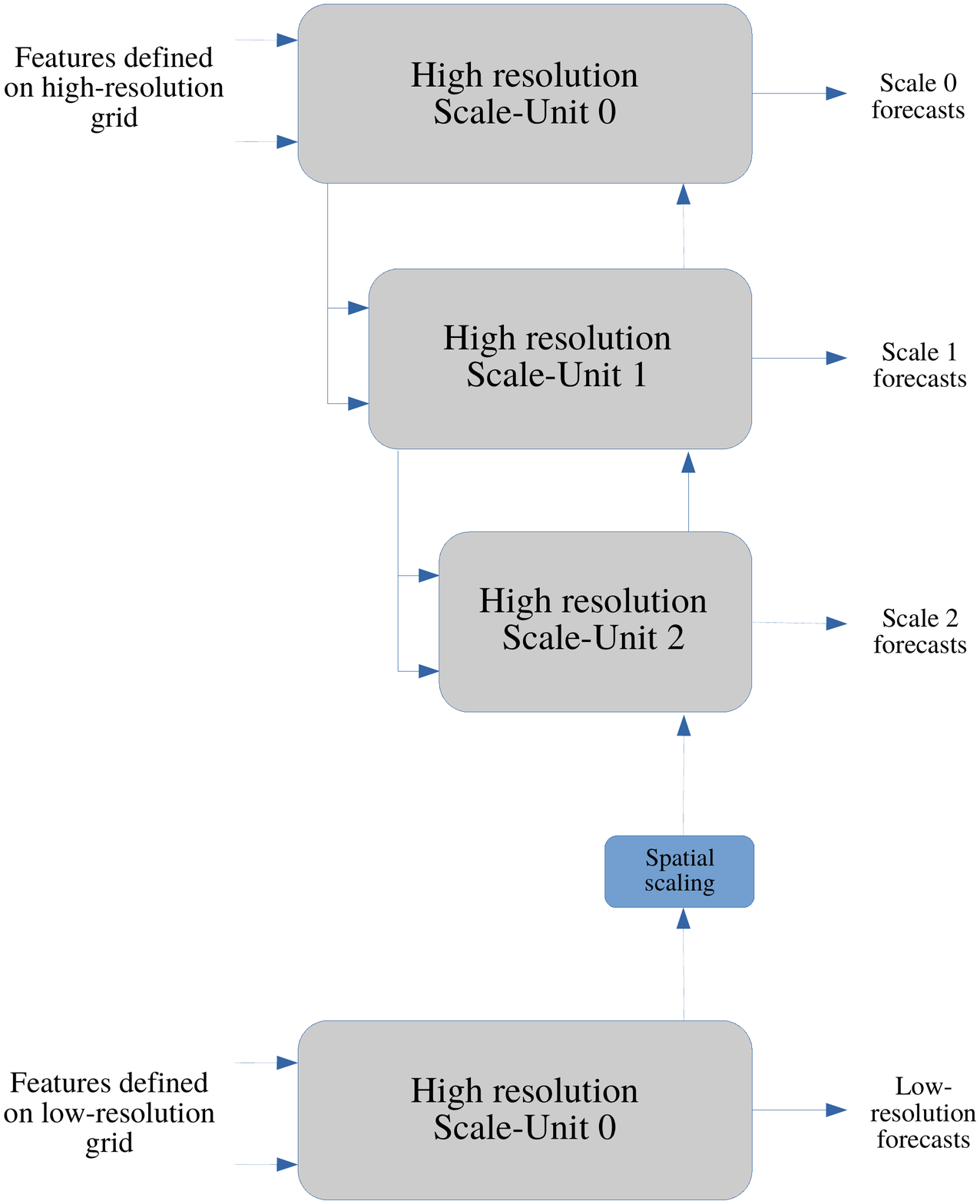}
  \caption{Model's architecture}
  \label{fig:schema_model}
\end{figure}

The engine described in this paper uses two input resolutions and four output resolutions and can be summarized as two main blocks, as schematized on Figure~\ref{fig:schema_model}:
\begin{itemize}
  \item The high-resolution block is formed with three stacked \textit{Scale-Unit} blocks.
The first one is fed with features projected on the high-resolution grid and outputs air quality forecasts at this same resolution (the highest one), and the two other ones output forecasts with lower resolutions ($100$ m and $200$ m respectively)
  \item The low-resolution block inputs features projected on the low-resolution grid and outputs air quality forecasts at this same resolution
The decoded features it produces are fed to the bottom part of the high-resolution block.
A layer named \textit{Spatial Scaling} is needed to project those decoded features on the other block.
They are first cropped to build the smallest grid that contains the spatial support of the high-resolution grid, and then spatially averaged and replicated
\end{itemize}

The \textit{Scale-Unit} blocks are parameterized the following way:
\begin{itemize}
  \item The encoders for historical features are composed of a TimeDistributedConv2D layer followed by a ConvLSTM layer
  \item The encoders for constant features are composed of a single TimeDistributedConv2D layer
  \item The encoders for forecast features are composed of two stacked ConvLSTM layers
  \item The decoder is made of a single TimeDistributedConv2D layer
\end{itemize}

Table~\ref{tab:hyper-parametrisation} lists the number of filters used in all the convolutional layers defined in the network.
\begin{table*}
  \caption{Number of filters in all convolutional layers}
  \label{tab:hyper-parametrisation}
  \begin{tabular}{c||c|cccc}
    \hline
    Grid & \textit{Scale-Unit} number & Historical encoder & Constant encoder & Forecast encoder & Decoder \\
    \hline
    \multirow{3}{*}{High-resolution} & 0 & 8, 16 & 1 & - & 8 \\
    & 1 & 8, 16 & 1 & - & 32 \\
    & 2 & 8, 16 & 1 & - & 64 \\
    \hline
    Low-resolution & 0 & 16, 32 & - & 64, 32 & 64 \\
    \hline
  \end{tabular}
\end{table*}

\subsection{Training of the engine}
\label{subsec:datasets'-description}

Let $\| l - l' \|$ be the euclidean distance between two locations $l$ and $l'$ and $k_d(l, l') = \exp\left(-\frac{\| l - l' \|^2}{2\sigma^2}\right)$ be the gaussian kernel, where the distance $\sigma$ is expressed in meters in the local two-dimensional euclidean frame.
The gaussian kernel is used afterwards to project on a grid the available monitoring stations' measurements around.
More precisely, each grid cell contains the weighted average of the measurements, and the weights corresponding to each station are computed with a gaussian kernel $k_d$.

The training of the engine is performed on patches of sizes $(64, 64)$ and $(20, 20)$ for the high-resolution and low-resolution grids respectively.
This means that those patches cover squares of length $3.2$ and $400$ kilometers respectively.

A patch is formed with the following grids, gathering input features and the targets needed to compute the training loss:

\subsubsection*{Features projected on the high-resolution grid}

\begin{itemize}
  \item Historical air quality monitoring stations' measurements: for each hour in the $N_{in}$ historical time steps, each grid cell contains a weighted average of the measurements computed with a gaussian kernel with $\sigma=5$ kilometers
  \item Historical road traffic estimates: for each hour in the $N_{in}$ historical time steps, a grid cell contains the value $\textit{Traffic}$ as defined in Section~\ref{sec:traffic}
  \item Road networks: a grid cell contains the value $\textit{RoadNetwork}$ as defined in Section~\ref{sec:roads}
\end{itemize}

\subsubsection*{Features projected on the low-resolution grid}

\begin{itemize}
  \item Historical air quality monitoring stations measurements: for each hour in the $N_{in}$ historical time steps, each grid cell contains a weighted average of the measurements computed with a gaussian kernel with $\sigma=50$ kilometers
  \item Weather forecasts and physical models' outputs: for each hour in the $N_{out}$ forecast time steps, a grid with the weather forecasts and physical models' outputs is built.
  Those forecasts are produced on a different grid than the grid used by the engine, hence they are first projected into the local euclidean frame and then interpolated using a simple bilinear interpolation
\end{itemize}

It is worth noting that stations' measurements are processed with both grids.
We have noticed that it was the best performing option: projections on the high-resolution grid enable to keep very localized information where the stations network is dense enough, while projections on the low-resolution grid enable to cover a larger area.
This choice enables to capture the phenomena affecting air quality at different scales.

\subsubsection*{Targets}

For every resolution $R$ output by the model (i.e. $R=50$ m, $R=100$ m, $R=200$ m and $R=20$ km) and for every hour in the $N_{out}$ forecast time steps, a grid cell of the target data is a weighted average of the available measurements around, computed with the gaussian kernel $k_d$ with $\sigma=3R$.
All cells whose weights' sum is under a threshold $T=\frac{1}{2}$ are set to $NaN$ and are ignored from the computation of the training loss.
This ensures that the training is uniquely performed on cells where there are enough monitoring stations around.

To make sure that the patches included in the training dataset contain at least one monitoring station, we have chosen to center each patch on a monitoring station.
Hence, a patch is defined by a starting time step (from January 1st, 2020 to December 31st, 2020) and a monitoring station on which all grids are centered.
To avoid spatial over-fitting, all measurements coming from the station on which the grids are centered are ignored to build the grids of a patch.

The model has been implemented with Keras and Tensorflow and trained on an Nvidia Geforce RTX $3090$.
The loss used in training is the mean squared logarithmic error (MSLE) loss.
Adam optimizer is used with a learning rate equal to $0.001$.
The batch size is $48$.
The number of epochs is $20$.

\section{Evaluation of the forecasting engine}
\label{sec:evaluation-of-the-forecasting-engine}

This section provides a detailed analysis of the accuracy of the forecasts produced by the engine in Europe and the United States.
The engine has been trained as described in the previous section on the $100$ cities with the most monitoring stations in Europe and in the United States.
$80\%$ of the cities are picked randomly to form the training dataset (containing $5.2$ million patches), and the remaining $20\%$ form the evaluation dataset ($1.2$ million patches).

\subsection{Forecasts' accuracy}
\label{subsec:forecasts'-accuracy}

The forecasts are evaluated at the locations of the monitoring stations where the reported measurements are the targets against which the forecasts are evaluated.
To be able to evaluate the engine in real conditions, the measurement reported at the station is not used in the prediction engine and the benchmarks.
We define the $2$ following benchmarks against which the prediction engine is evaluated:
\begin{itemize}
  \item \textit{Closest measurement}: constant forecast equal to the measurement returned at time $t=0$ by the closest monitoring station
  \item \textit{Advanced benchmark}: let $(x, y)$ be the location at which the model is evaluated.
  We introduce $DeepPlume_{0, x, y}$ the prediction obtained at time $t=0$ by the spatial prediction model introduced in ~\cite{deepplume}, and $PlumeNet_{t, x, y}$ be the prediction obtained at time $t$ by the spatiotemporal prediction model introduced in ~\cite{plumenet}.
  Then, the advanced benchmark $B_{t, x, y}$ at time $t$ is defined as $B_{t,x,y}=DeepPlume_{0,x,y} + PlumeNet_{t,x,y}-PlumeNet_{0,x,y}$.
  This benchmark combines the very good accuracy and granularity of spatial predictions by the DeepPlume model introduced in~\cite{deepplume}, and the forecasting ability demonstrated by the PlumeNet model introduced in~\cite{plumenet}
\end{itemize}

\begin{figure}[h]
  \centering
  \includegraphics[width=\linewidth]{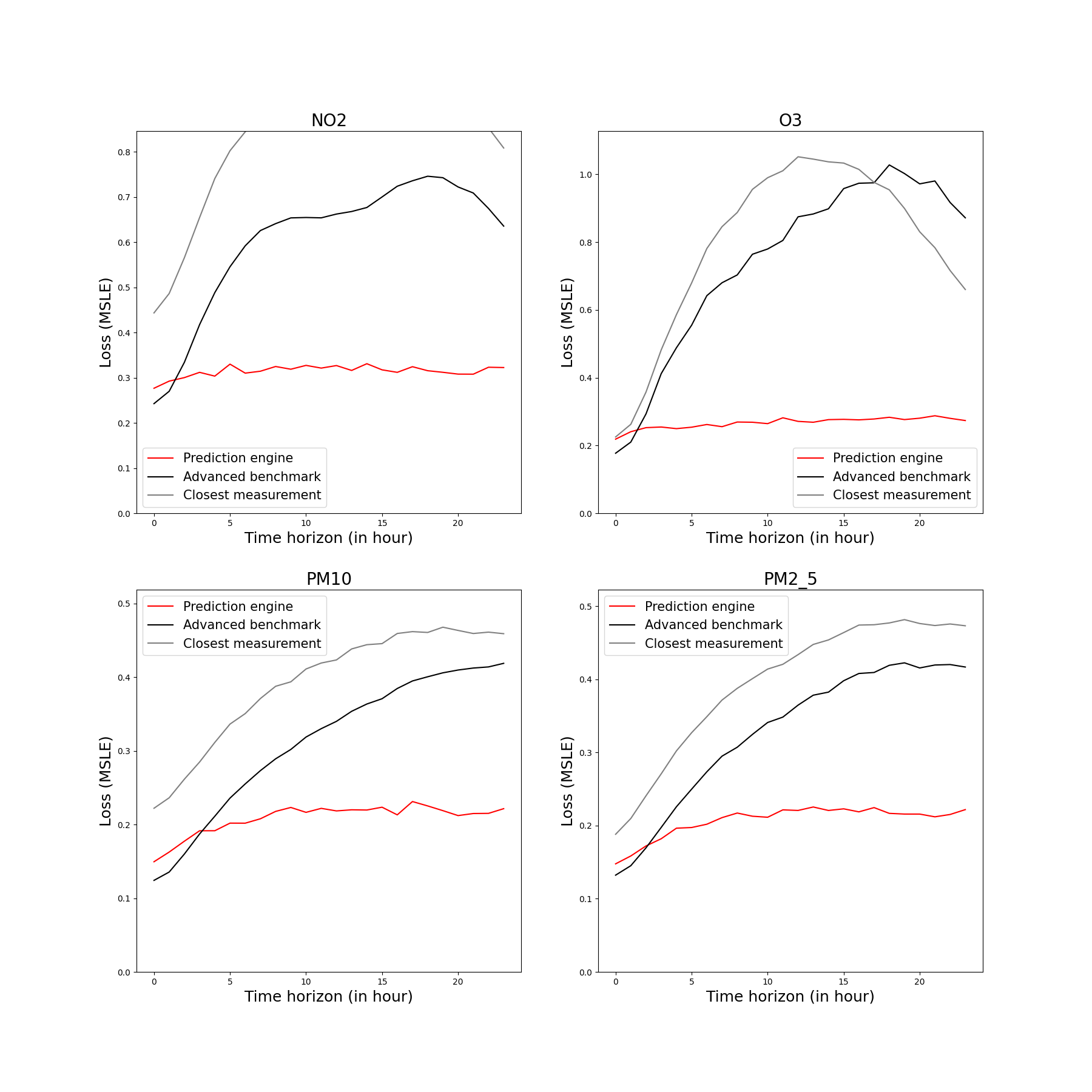}
  \caption{Loss as a function of the time horizon}
  \label{fig:time_horizon_loss}
\end{figure}

Tables~\ref{tab:accuracy} and ~\ref{tab:accuracy_us} give the evaluation loss MSLE averaged until $24$ hours in Europe and the United States respectively.
The prediction model gives a very significant improvement compared to the benchmarks for all pollutants.
In Figure~\ref{fig:time_horizon_loss}, we notice that the advanced benchmark outperforms our prediction model in the first few forecast time steps.
This can be explained by the fact that in the first time steps, the advanced benchmark is mainly based on the \textit{DeepPlume} model, which focuses on spatial prediction and is fed with more inputs than our model (for example land-use data).
However, our prediction model is much more powerful to produce accurate forecasts after that.

\begin{table}[h!]
  \caption{Loss averaged over the time horizon (Europe)}
  \label{tab:accuracy}
  \begin{tabular}{c|ccc}
    \hline
     & \multicolumn{3}{c}{24 hours} \\
     & Closest & Advanced & Prediction \\
     & measurement & benchmark & engine \\
    \hline
    NO2 & 0.831 & 0.604 & 0.314 \\
    O3 & 0.794 & 0.743 & 0.266 \\
    PM10 & 0.393 & 0.312 & 0.208 \\
    PM2.5 & 0.395 & 0.327 & 0.206 \\
    \hline
    Global & 0.603 & 0.497 & 0.249 \\
    \hline
\end{tabular}
\end{table}

\begin{table}[h!]
  \caption{Loss averaged over the time horizon (United States)}
  \label{tab:accuracy_us}
  \begin{tabular}{c|ccc}
    \hline
     & \multicolumn{3}{c}{24 hours} \\
     & Closest & Advanced & Prediction \\
     & measurement & benchmark & engine \\
    \hline
    NO2 & 0.794 & 0.564 & 0.367 \\
    O3 & 0.799 & 0.589 & 0.243 \\
    PM10 & 0.542 & 0.428 & 0.280 \\
    PM2.5 & 0.442 & 0.380 & 0.277 \\
    \hline
    Global & 0.644 & 0.490 & 0.292 \\
    \hline
\end{tabular}
\end{table}

\subsection{Illustration}
\label{subsec:illustrations}

\begin{figure*}[h]
  \centering
  \includegraphics[width=.21\linewidth]{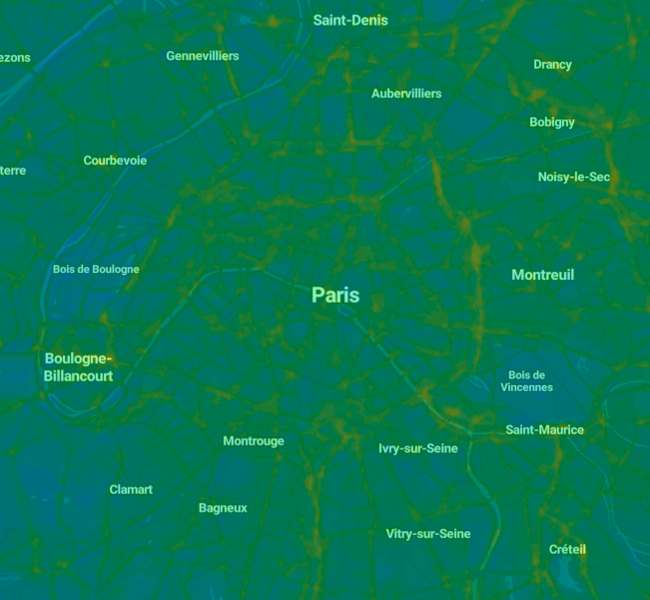}
  \includegraphics[width=.21\linewidth]{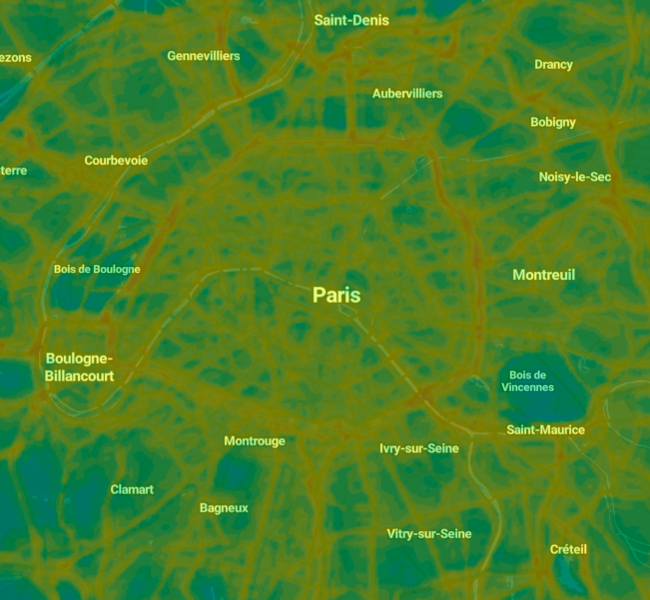}
  \includegraphics[width=.21\linewidth]{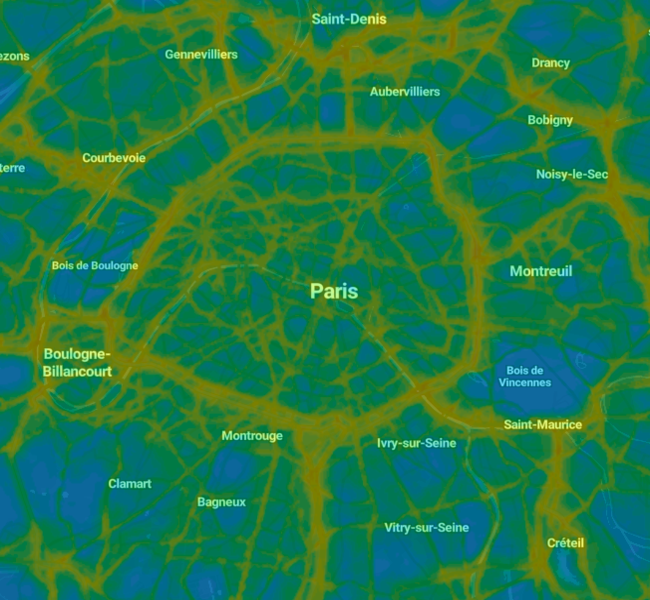}
  \includegraphics[width=.21\linewidth]{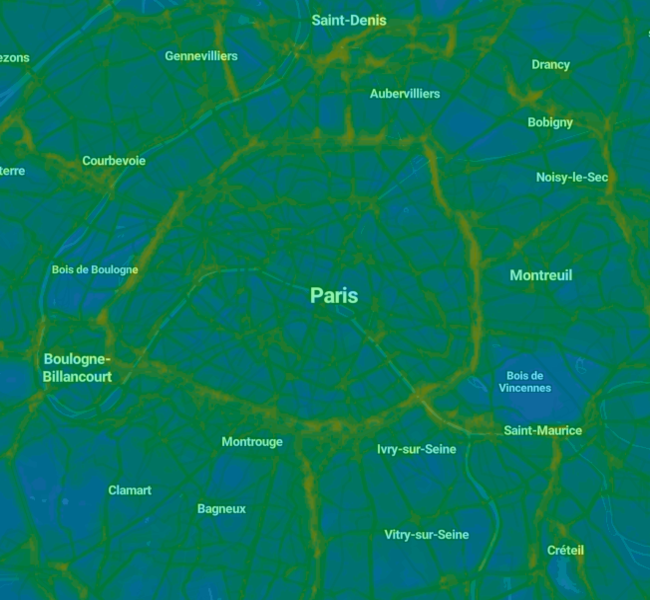} \\
  \includegraphics[width=.21\linewidth]{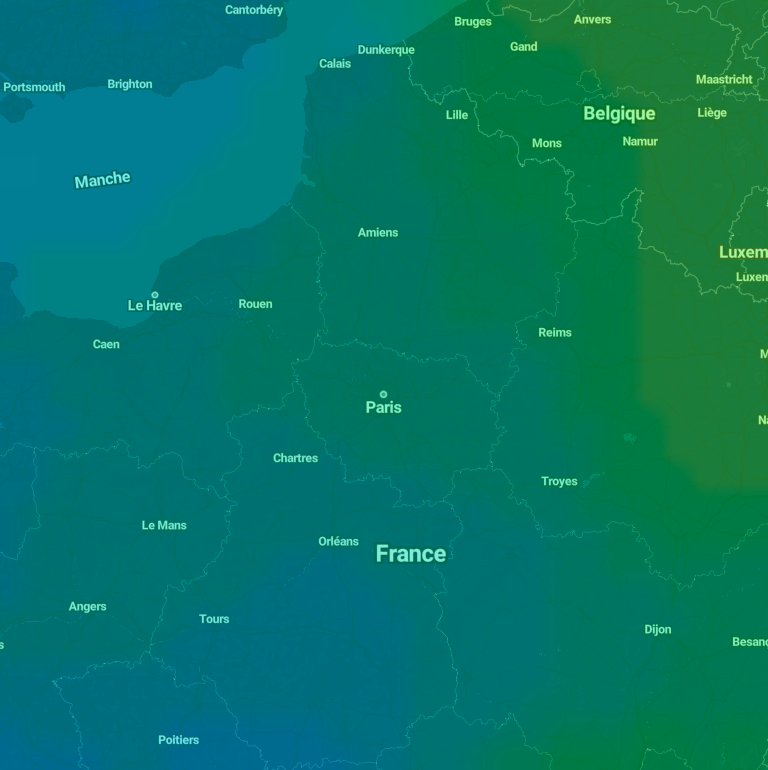}
  \includegraphics[width=.21\linewidth]{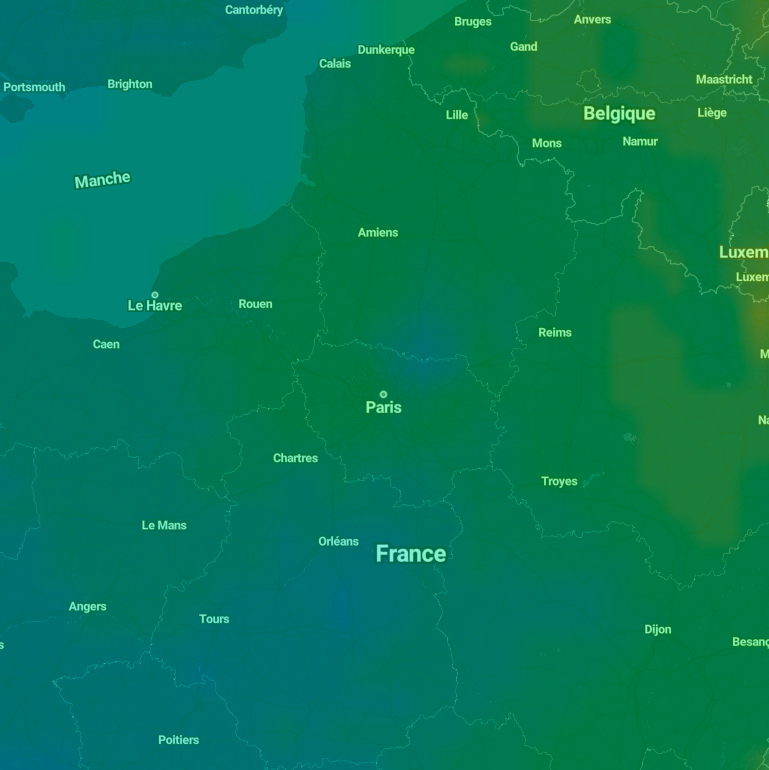}
  \includegraphics[width=.21\linewidth]{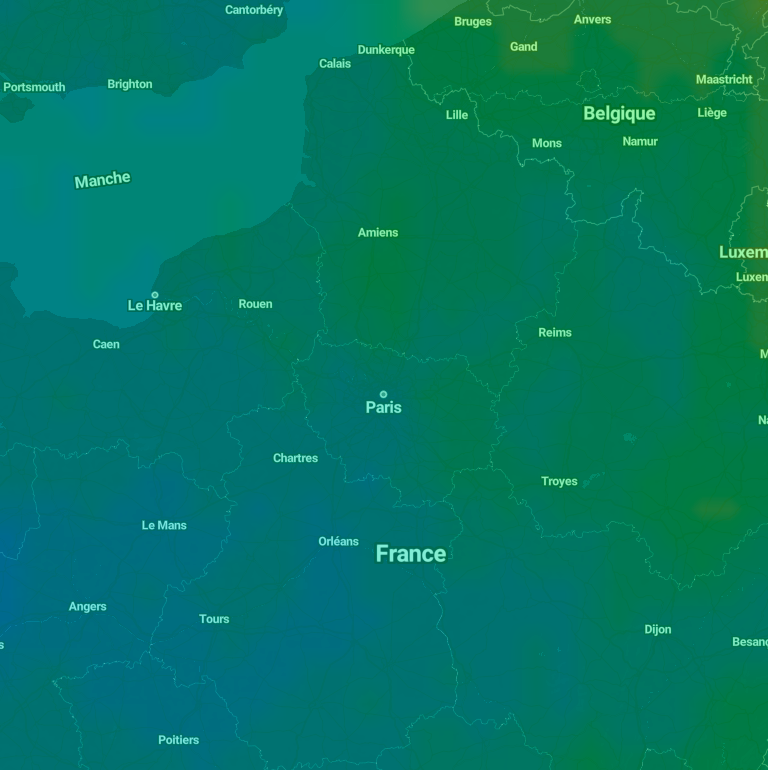}
  \includegraphics[width=.21\linewidth]{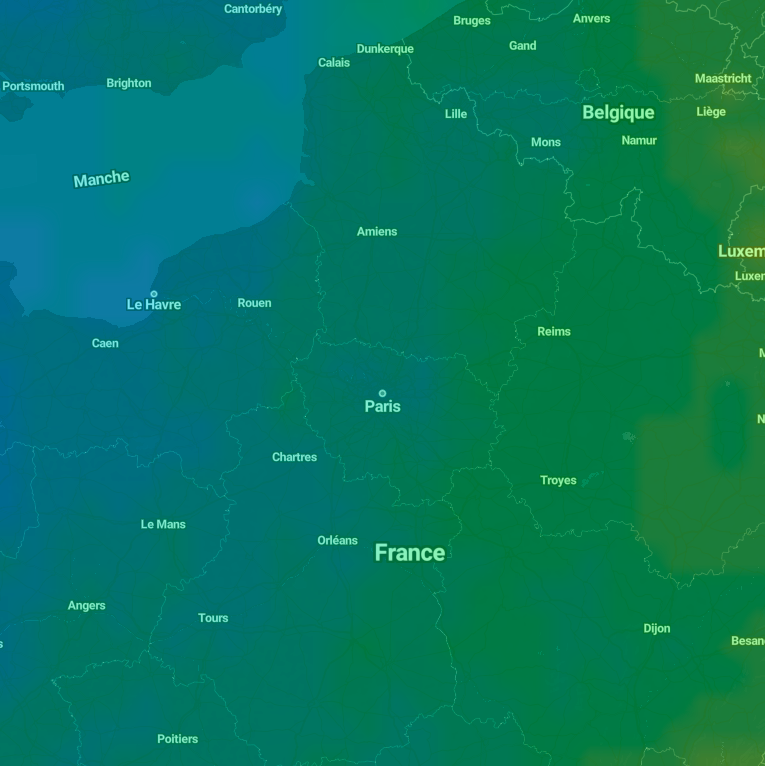} \\
  \includegraphics[width=.40\linewidth]{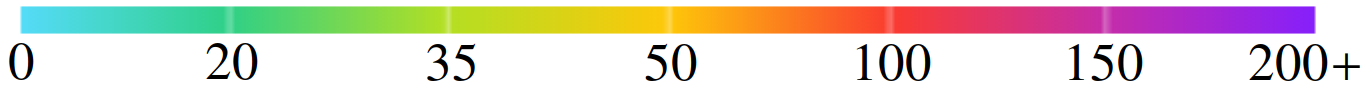}
  \caption{NO$_2$ forecasts at $50$ meters (first row) and $20$ kilometers (second row) for the $0, 6, 12 \text{ and } 18$ hour horizons  (Paris, 11/11/2020)}
  \label{fig:no2_forecasts}
\end{figure*}

As an illustration, Figure~\ref{fig:no2_forecasts} shows NO$_2$ forecasts produced in Paris on 11/11/2020.
The forecasts are presented on $2$ spatial resolutions: $50$ meters and $20$ kilometers, which are the highest and lowest resolutions produced by the model.
The concentrations are expressed using Plume Labs' internal Air Quality Index\footnote{An Air Quality Index~(AQIs) is a normalization of a pollutant raw concentration in $\mu g/m^{3}$ to a health impact scale, allowing inter-pollutant comparison.
AQIs are commonly used worldwide and are usually defined locally at the country or continent scale to comply with local standards.
Plume Labs AQI is based on WHO recommendations and is extensively described here: \url{https://plumelabs.zendesk.com/hc/en-us/articles/360008268434-What-is-the-Plume-AQI-}.}.

\section{Conclusion and future work}
\label{sec:conclusion-and-future-work}

The prediction engine presented in this paper produces air quality forecasts at different spatial resolutions, from a few dozens of meters to dozens of kilometers, making it relevant for various use-cases fed with air quality data.
It is based on a U-Net architecture formed with \textit{Scale-Unit} blocks, able to integrate seamlessly the different inputs available: air quality monitoring stations' measurements, traffic data, weather forecasts and physical models' outputs.
We have shown in this paper that it performs much better than simpler methods which fail to properly integrate the spatiotemporal patterns happening at different scales.
An important characteristic of the engine is that it keeps a good out-of-sample accuracy, meaning that it is able to generalize beyond its training dataset and hence can be applied to cities with few monitoring stations, which have not been included in this study.

Another very valuable advantage of this engine compared to most existing physical models is that it needs much less computing power.
As an illustration, producing $24$-hour forecasts around a city at a given time takes a few minutes with a standard CPU, compared to a few hour for traditional approaches.
Hence, the engine's forecasts can be updated very frequently and integrate the most recent air quality monitoring stations' measurements and traffic estimates.

We think that the engine can still be improved in several ways.
First, we focused in this paper on producing $24$-hour forecasts, but it is straightforward to use a similar architecture for longer time horizons, at the cost of longer training and inference times.
Another important margin for improvement lies in the data sources fed to the model.
Additional data sources like land-use data or traffic forecasts could help to improve the forecasts' accuracy.
Finally, we have noticed that using a multi-resolution architecture helped to build robust prediction models able to generalize beyond the training dataset.
We can expect that a model based on a more complex architecture with more intermediate resolutions between the low and high resolutions used here should improve significantly the model's accuracy.


\bibliographystyle{ACM-Reference-Format}
\bibliography{temporal_prediction}










\end{document}